\documentclass[sigconf,nonacm]{acmart}

\usepackage{multirow}
\usepackage{enumitem}  
\usepackage{rotating}
\usepackage{placeins} 
\usepackage{float} 
\usepackage{float} 
\usepackage{cleveref} 

\AtBeginDocument{%
  }

\copyrightyear{2025}
\acmYear{2025}
\setcopyright{acmlicensed}\acmConference[MM '25]{Proceedings of the 33rd ACM International Conference on Multimedia}{October 27--31, 2025}{Dublin, Ireland}
\acmBooktitle{Proceedings of the 33rd ACM International Conference on Multimedia (MM '25), October 27--31, 2025, Dublin, Ireland}
\acmDOI{10.1145/3746027.3754908}
\acmISBN{979-8-4007-2035-2/2025/10}

\begin{document}

\title{RWKV-PCSSC: Exploring RWKV Model for Point Cloud Semantic Scene Completion}


\author{Wenzhe He}
\authornote{Both authors contributed equally to this research.}
\orcid{0009-0001-1093-1963}
\affiliation{%
  \institution{Hunan University}
  \department{College of Computer Science and Electronic Engineering}
  \city{Changsha}
  \state{Hunan}
  \country{China}
}
\email{hewenzhe@hnu.edu.cn}

\author{Xiaojun Chen}
\authornotemark[1]
\orcid{0009-0005-1440-4473}
\affiliation{%
  \institution{Hunan University}
  \department{College of Computer Science and Electronic Engineering}
  \city{Changsha}
  \state{Hunan}
  \country{China}
}
\email{chenxiaojun@hnu.edu.cn}

\author{Wentang Chen}
\affiliation{%
  \institution{Hunan University}
  \department{College of Computer Science and Electronic Engineering}
  \city{Changsha}
  \state{Hunan}
  \country{China}
}
\orcid{0009-0001-4418-1531}
\email{B241000657@hnu.edu.cn}

\author{Hongyu Wang}
\affiliation{%
  \institution{Hunan University}
  \department{College of Computer Science and Electronic Engineering}
  \city{Changsha}
  \state{Hunan}
  \country{China}
}
\orcid{0009-0003-5113-3735}
\email{wanghongyu@hnu.edu.cn}

\author{Ying Liu}
\authornote{Corresponding authors}
\affiliation{%
  \institution{Hunan Normal University}
  \department{College of Information Science and Engineering}
  \city{Changsha}
  \state{Hunan}
  \country{China}
}
\orcid{0000-0003-3740-9144}
\email{liu_ying@hunnu.edu.cn}

\author{Ruihui Li}
\authornotemark[2]
\affiliation{%
  \institution{Hunan University}
  \department{College of Computer Science and Electronic Engineering}
  \city{Changsha}
  \state{Hunan}
  \country{China}
}
\orcid{0000-0002-4266-6420}
\email{liruihui@hnu.edu.cn}

\renewcommand{\shortauthors}{Wenzhe He et al.}

%
\begin{abstract}
Semantic Scene Completion (SSC) aims to generate a complete semantic scene from an incomplete input. Existing approaches often employ dense network architectures with a high parameter count, leading to increased model complexity and resource demands. To address these limitations, we propose RWKV-PCSSC, a lightweight point cloud semantic scene completion network inspired by the Receptance Weighted Key Value (RWKV) mechanism.
Specifically, we introduce a RWKV Seed Generator (RWKV-SG) module that can aggregate features from a partial point cloud to produce a coarse point cloud with coarse features. Subsequently, the point-wise feature of the point cloud is progressively restored through multiple stages of the RWKV Point Deconvolution (RWKV-PD) modules. 
By leveraging a compact and efficient design, our method achieves a lightweight model representation. Experimental results demonstrate that RWKV-PCSSC reduces the parameter count by 4.18$\times$ and improves memory efficiency by 1.37$\times$ compared to state-of-the-art methods PointSSC\cite{yan2024pointssc}. 
Furthermore, our network achieves state-of-the-art performance on established indoor (SSC-PC, NYUCAD-PC) and outdoor (PointSSC) scene dataset, as well as on our proposed datasets (NYUCAD-PC-V2, 3D-FRONT-PC).
\end{abstract}






\maketitle

\section{Introduction}

Semantic scene completion (SSC) aims to generate a complete point cloud with accurate semantic labels from a partial point cloud input. SSC finds applications in domains such as autonomous driving and robotic navigation \cite{gupta2017cognitive,varley2017shape,liang2021sscnav}. However, sensor limitations and environmental complexities often lead to highly incomplete and noisy input data, posing significant challenges for scene completion and semantic interpretation.

\begin{figure*}[t!]
  \centering
  \includegraphics[width=\textwidth]{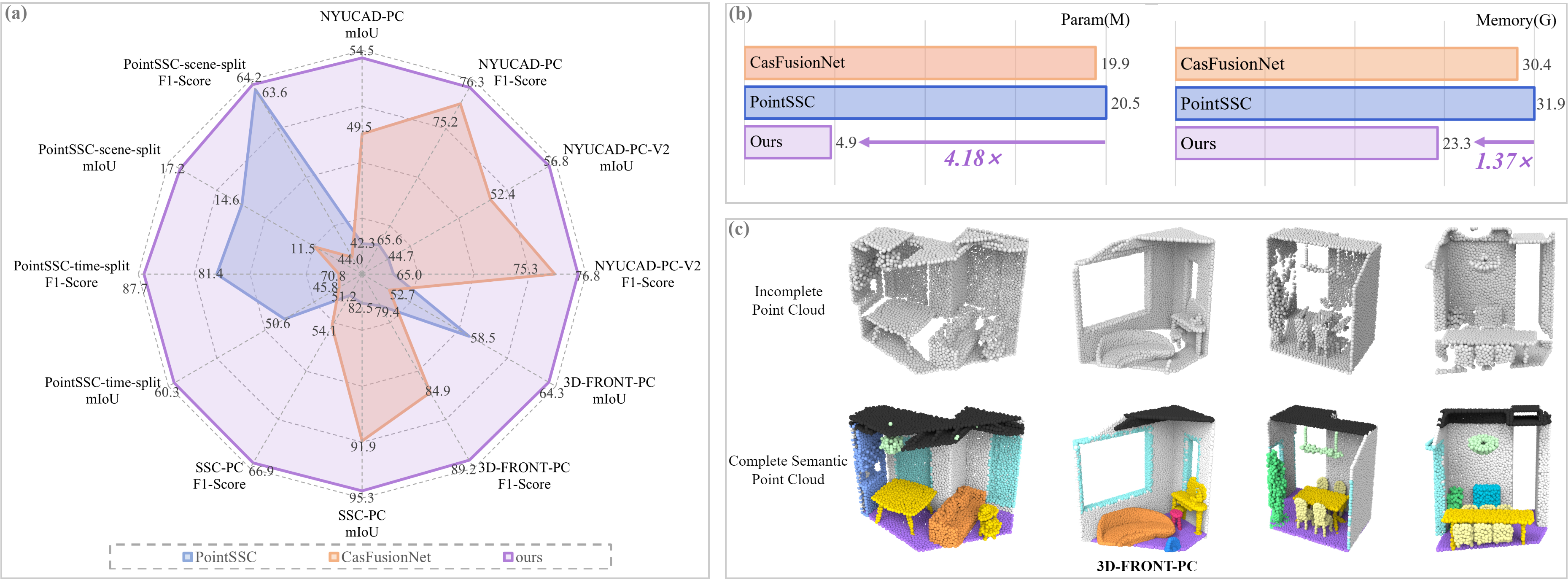}
  \caption{(a) Compared to CasFusionNet~\cite{Xu2023CasFusionNet} and PointSSC~\cite{yan2024pointssc}, RWKV-PCSSC achieves state-of-the-art performance in both scene completion and segmentation across a variety of indoor and outdoor datasets.  
(b) On PointSSC datasets, RWKV-PCSSC significantly reduces parameters and memory usage while maintaining competitive performance.  
(c) We introduce \textbf{3D-FRONT-PC}, a novel semantic scene completion dataset derived from 3D-FRONT~\cite{fu20213d3d-front,fu20213d3d-future} scene models. Compared to NYUCAD-PC, 3D-FRONT-PC presents more challenging tasks (i.e., retaining input point cloud coordinates in the world coordinate system), covers more categories (i.e., increased from 11 to 20), and scales significantly larger (i.e., expanded from 1,449 to 12,655 instances).}
  \Description{Enjoying the baseball game from the third-base
  seats. Ichiro Suzuki preparing to bat.}
  \label{fig:teaser}
\end{figure*}

SSCNet \cite{song2017semantic} first proposed the use of voxel-based methods to address the SSC task, and since then, many subsequent works \cite{gupta2017cognitive,varley2017shape,liang2021sscnav,Li2021Anisotropic,Liu20242D,Wang2024Multi-modal} have adopted voxel-based approaches for SSC. Voxel representation necessitates the discretization of 3D space, resulting in the generation of a large number of voxel units. As resolution increases, the computational and storage demands also increase significantly, leading to high computational costs. 

Considering the limitations of voxel representation \cite{zhang2021point,Xu2023CasFusionNet}, methods such as PCSSC-Net \cite{zhang2021point} have been proposed to address these challenges. Specifically, PCSSC-Net uses a 12-dimensional feature vector, which includes point coordinates, RGB values, normal vectors, and differential coordinates, as input to the network to generate a complete segmented point cloud. Although PCSSC-Net has yielded promising results, they do not adequately account for the important relationship between scene completion and semantic segmentation. CasFusionNet \cite{Xu2023CasFusionNet} aims to achieve dense feature fusion between the two tasks, demonstrating commendable performance. However, it directly downsamples the input to generate coarse point clouds, which oversimplifies the task, fails to predict missing regions, and loses geometric information. 
PointSSC \cite{yan2024pointssc} has demonstrated promising performance on its proposed outdoor scene dataset. However, it fails to capture the point cloud's intricate features fully. Furthermore, both PointSSC and CasFusionNet suffer from complex architectures, resulting in a large number of parameters and significant memory usage.

The latest advancements in the RWKV model for Natural Language Processing (NLP) demonstrate its efficiency in handling global information and sparse inputs. Building on these strengths, we propose a novel method for semantic scene completion, termed RWKV-PCSSC, which initially integrates an RWKV seed generator (RWKV-SG) module to construct a coarse point cloud enriched with point-wise features. These features are subsequently refined through multiple RWKV Point Deconvolution (RWKV-PD) module stages. RWKV-PD enables iterative refinement and segmentation of the coarse point cloud to progressively achieve precise scene completion and segmentation.

Currently, research on point cloud semantic scene completion (PCSSC) remains relatively limited, partly due to the scarcity of relevant datasets. Therefore, To address this shortfall, we introduce 3D-FRONT-PC, a novel dataset derived from 3D-FRONT scene models. Compared to the NYUCAD-PC dataset, 3D-FRONT-PC presents three major challenges. First, it requires preserving input point cloud coordinates in the world coordinate system. Second, the category range is expanded from 11 to 20 classes. Third, the dataset size is significantly increased from 1,449 to 12,655 instances. Additionally, we propose NYUCAD-PC-V2, an enhanced iteration of NYUCAD-PC that resolves semantic ambiguities present in the original dataset.
To validate the effetiveness of our method, we evaluated RWKV-PCSSC on four indoor datasets: 3D-FRONT-PC, NYUCAD-PC-V2, NYUCAD-PC \cite{firman2016structured,Xu2023CasFusionNet}, and SSC-PC \cite{zhang2021point}, as well as on the outdoor dataset PointSSC \cite{yan2024pointssc} with its two data division strategies, split by time and split by scene.
Our network performed excellently well in quantitative and qualitative experiments, significantly outperforming existing state-of-the-art methods. \Cref{fig:teaser} (a) compares our method with others in terms of segmentation and completion performance. In the indoor datasets SSC-PC \cite{liang2021sscnav}, NYUCAD-PC \cite{Xu2023CasFusionNet}, NYUCAD-PC-V2, and 3D-FRONT-PC, as well as the outdoor datasets PointSSC-scene-split and PointSSC-time-split \cite{yan2024pointssc}, our approach consistently achieves state-of-the-art performance in both segmentation and completion metrics. Furthermore, \Cref{fig:teaser} (b) compared to the state-of-the-art method \cite{yan2024pointssc} on PointSSC \cite{yan2024pointssc} dataset, our approach achieves a 76.1\% reduction in parameter count and a 27.0\% reduction in memory usage.  This research not only provides a new perspective for semantic scene completion but also offers robust support for related fields of study.

In conclusion, the contributions of our paper can be summarized as follows: 
\begin{itemize}
    \item RWKV-PCSSC achieves state-of-the-art performance in point cloud semantic scene completion across diverse datasets, outperforming PointSSC \cite{yan2024pointssc} with 76.1\% fewer parameters and 27.0\% less memory.
    \item We designed two novel modules: the RWKV seed generator (RWKV-SG) module to generate an initial coarse point cloud with point-wise features, which is iteratively refined and segmented through the RWKV Point Deconvolution (RWKV-PD) module for precise point cloud completion and segmentation.
    \item We present two datasets: 3D-FRONT-PC, which comprises 12,655 point cloud scene with 20 label categories derived from the 3D-FRONT\cite{fu20213d3d-future,fu20213d3d-front} dataset, and NYUCAD-PC-V2, an enhanced version of NYUCAD-PC\cite{firman2016structured,Xu2023CasFusionNet} that addresses semantic ambiguities in the original dataset.
\end{itemize}

\section{Related Work}
\label{sec:Related_Work}

\textbf{Point Cloud Segmentation.} PointNet \cite{Qi2016PointNet} and PointNet++ \cite{qi2017pointnet++} pioneered point-based networks using Set Abstraction to model point relationships. Subsequent works \cite{Wang2018Dynamic,Chenfeng2020SqueezeSegV3,thomas2019KPConv,Te2018RGCNN,Qian2022PointNeXt,Ma2022Rethinking} advanced segmentation by addressing fine-grained geometric and contextual challenges in large-scale 3D environments. Attention-based methods \cite{zhao2021ptv1,Chenfeng2020SqueezeSegV3,Wu2022Point,Lai2022Stratified,wu2024ptv3,Guo2021PCT,Thomas2024KPConvX} demonstrated superior performance by effectively handling the permutational invariance of point clouds. Mamba-based approaches \cite{liang2024pointmamba,Han2024Mamba3D,zhang2024point} have shown improved scalability and efficiency compared to transformers.

\textbf{Point Cloud Completion.} PCN \cite{yuan2018pcn} pioneered completion by learning latent shape representations to generate missing points. Subsequent methods \cite{Wu2018PointConv,liu2020morphing,zong2021ashf,Yu2023AdaPoinTr,Rong2024CRA-PCN,Tchapmi2019TopNet,zhou2022seedformer} have been proposed to advance point cloud completion, significantly improving robustness and accuracy. SnowflakeNet \cite{xiang2022snowflake} introduced a snowflake-like growth process for point generation. PointAttN \cite{Wang2024PointAttN} employs attention mechanisms to model local geometric relationships, demonstrating significant performance.

\textbf{Receptance Weighted Key Value (RWKV) \cite{peng2023rwkv}.} The Receptance Weighted Key Value (RWKV) model integrates the parallel training efficiency of transformers with the linear computational scalability of RNNs, offering a compelling solution for sequence processing tasks. By utilizing a redesigned linear attention mechanism, RWKV overcomes the quadratic scaling limitations inherent in traditional Transformers, making it particularly well-suited for long-sequence reasoning. Furthermore, its successful adaptation to visual domains \cite{wang2024occrwkv,Yang2024Restore-RWKV,Duan2024Vision-RWKV,Fei2024Diffusion-RWKV,he2024pointrwkv}, exemplifies its versatility. for instance, Vision-RWKV \cite{Duan2024Vision-RWKV} has been implemented for high-resolution image processing
, while PointRWKV \cite{he2024pointrwkv} facilitates effective 3D point cloud encoding
. These adaptations underscore the model’s efficiency and broad applicability in complex tasks, including advanced 3D spatial analysis.

\textbf{Semantic Scene Completion.} Existing segmentation and completion methods often neglect the interdependencies between these tasks, leading to suboptimal results due to their inherent complementarity. Semantic scene completion (SSC) has emerged as a pivotal task in 3D computer vision, aiming to generate complete and semantically annotated 3D scenes from partial observations. Recent studies \cite{Chen20203D,Li2021Anisotropic,Wang2023Semantic,Liu20242D,Wang2024Multi-modal} predominantly adopt voxel-based representations, discretizing incomplete scenes into 3D voxel grids. These grids are processed using 3D Convolutional Neural Networks (3D CNNs), which excel at capturing spatial context and local dependencies. For example, SSCNet \cite{song2017semantic} pioneered SSC by employing an end-to-end 3D convolutional network. Subsequent works \cite{zhang2018efficient,li2020anisotropic} improved voxel-based modeling through modified 3D convolutions, while \cite{Li2023DDIT} introduced instance-level completion by deforming templates with latent codes. Additionally, recent advancements \cite{dourado2021edgenet,dourado2022data,Liu20242D} integrate depth and color features into 2D semantic networks, further enhancing performance.

\textbf{Point Cloud Semantic Scene Completion.} Although voxel-based methods enable effective volumetric data processing, they often incur high computational costs due to excessive parameters and voxel grid memory overhead. In contrast, point clouds provide a more memory-efficient and flexible representation of 3D data, capturing intricate geometric details while avoiding the resolution limitations inherent in voxel-based methods. SSCPC-Net \cite{zhang2021point} addresses semantic scene completion using a 12-dimensional feature vector as input and designing a patch-based contextual encoder to predict a completed and segmented point cloud. Recently, CasFusionNet \cite{Xu2023CasFusionNet} attempted to bridge this gap by promoting feature exchange between the completion and segmentation modules, thus improving the interaction between these tasks. PointSSC \cite{yan2024pointssc} achieves efficient and accurate semantic scene completion by effectively integrating global and local features for large-scale outdoor environments. However, these methods are often limited by their complex and cumbersome architecture, leading to an overabundance of parameters and considerable memory usage, which hinders their practical scalability. Moreover, they demonstrate constrained capability in effectively capturing both geometric structures and semantic correlations, resulting in suboptimal performance for semantic scene completion tasks. In this paper, we propose RWKV-PCSSC, a novel point cloud semantic scene completion network that leverages the RWKV mechanism to enhance parameter efficiency and accuracy.

\section{Method}
\label{sec:Method}

\begin{figure*}[t!]
  \centering
  \includegraphics[width=\textwidth]{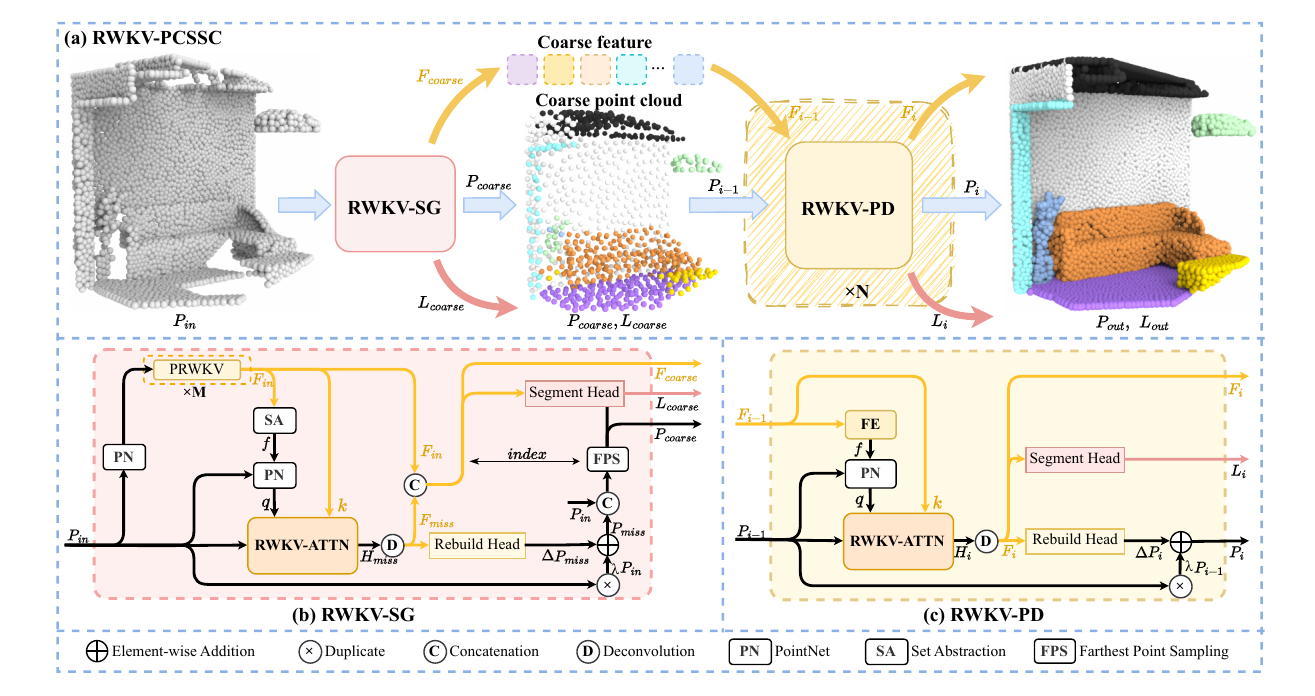}
  \caption{(a) The architecture of the RWKV-PCSSC network is depicted as follows. The network comprises an RWKV Seed Generator (RWKV-SG) module and a sequence of N stages, each consisting of RWKV Point Deconvolution (RWKV-PD) modules. (b) Detailed description of the RWKV-SG module. (c) The details of the RWKV-PD module.}
  \label{fig:PipLine}
\end{figure*}

\subsection{Overview Pipeline}
Given an incomplete point cloud \( P_{in} \in \mathbb{R}^{N \times 3} \) containing \( N \) points as input, our goal is to produce a complete 3D scene \( P_{out} \in \mathbb{R}^{M \times 3} \) with \( M \) points (\( M \geq N \)), along with the predicted per-point semantic labels \( L_{out} \in \mathbb{R}^{M \times C} \), where \( C \) denotes the total number of semantic classes. \Cref{fig:PipLine} (a)  illustrates the overall structure of RWKV-SSCPC: The network first produces coarse features alongside a semantically labeled coarse point cloud via the RWKV‑SG module, followed by multiple stages of RWKV-PD modules that refine the output in a coarse-to-fine manner \cite{wen2021pmp,wen2022pmp,zhou2022seedformer,Rong2024CRA-PCN,Wang2024PointAttN} to produce a completed and segmented point cloud. Below, we elaborate on the steps in our pipeline.

First, we input \( P_{in} \) into the RWKV-SG module, which generates a coarse point cloud with semantic labels, denoted as \( P_{coarse} \) and \( L_{coarse} \). Relying solely on \( P_{coarse} \) as inputs for the next stage of the RWKV-PD module can lead to the loss of details within the input point cloud data, which would negatively impact the overall refinement process. To fully leverage the features of the input point cloud, our RWKV-SG module also outputs the point-wise feature \( F_{coarse} \) of the coarse point cloud.
We then use \( P_{coarse} \) and \( F_{coarse} \) as inputs to the first RWKV-PD module. Specifically, \( P_{coarse} \) and \( F_{coarse} \) correspond to \( P_{i-1} \), \( F_{i-1} \) of the first RWKV-PD module, respectively. 
Throughout the \( N \) stages of refinement, for example, in the \( i \)-th stage (\( 2 \leq i \leq N \)), the RWKV-PD inputs consist of the outputs from the previous RWKV-PD stage. In the initial RWKV-PD stage (\( i = 1 \)), the input to RWKV-PD is sourced from the output of the preceding RWKV-SG. This hierarchical coarse-to-fine approach progressively reconstructs the point-wise features of the point cloud, ultimately yielding a fully complete segmented point cloud \( P_{out} \) and \( L_{out} \).

\subsection{RWKV Seed Generator}

Although several studies have addressed semantic scene completion tasks in point clouds, existing methods remain limited in their ability to predict detailed geometries for missing regions and effectively capture local structural information during coarse point cloud generation. To overcome these limitations, the RWKV Seed Generator (RWKV-SG) module was developed. This module extracts essential local and global geometric features from the input point cloud to predict and reconstruct missing regions, generates a coarse point cloud \(P_{coarse}\) with corresponding semantic labels \(L_{coarse}\), and produces detailed point-wise features \(F_{coarse}\), thereby furnishing adequate information for subsequent RWKV-PD modules. \Cref{fig:PipLine} (b) illustrates the detailed architecture of RWKV-SG.

We first employ the PointNet module \cite{qi2017pointnet++} to extract preliminary features from the input point cloud \( P_{in} \). Subsequently, to further explore the deep-level features within the point cloud data, we introduce the PRWKV module. Leveraging its robust sequence modeling capabilities, the PRWKV module enhances and abstracts the preliminary features across multiple levels (i.e., \(M=4\) as illustrated in \Cref{fig:PipLine} (b)), thereby obtaining more discriminative high-order feature representations \( F_{in} \). Next, we utilize the Set Abstraction module \cite{qi2017pointnet++} to extract the global feature \( f \). The global feature \( f \) is then combined with \( P_{in} \) and fed into the PointNet module to obtain the query \( q_{in} \). Subsequently, \( F_{in} \) is treated as \( k_{in} \), \( P_{in} \), \( q_{in} \), \( k_{in} \) and input into the RWKV-ATTN module to predict hidden information about missing parts, denoted as \( H_{miss} \). Subsequently, \( H_{miss} \) is passed into the Deconvolution module to obtain \( F_{miss} \). To derive \( \Delta P \), we employ an MLP with a tanh activation function as our Rebuild Head. \( \Delta P \) is then added to \( P_{in} \) to obtain \( P_{miss} \). To generate the complete coarse point cloud \( P_{coarse} \), we apply farthest point sampling (FPS) to the concatenation of \( P_{in} \) and \( P_{miss} \), while simultaneously obtaining its corresponding feature \( F_{coarse} \). Finally, we process \( F_{coarse} \) using a Segment Head composed of KPConvD \cite{Thomas2024KPConvX}, PRWKV and MLP to obtain the semantic labels \( L_{coarse} \).

The final outputs of the RWKV-SG: \( P_{coarse} \), \( F_{coarse} \), are propagated as hierarchical inputs to the first stage RWKV-PD module for progressive refinement.

\subsection{RWKV Point Deconvolution}

The purpose of RWKV Point Deconvolution (RWKV-PD) is to progressively refine the coarse point cloud \( P_{coarse} \) into a complete point cloud with semantic labels. \Cref{fig:PipLine} (c) shows the detailed architecture of our designed RWKV-PD.

Notably, the input for the first stage of RWKV-PD(\( i = 1 \)) in the network is derived from the preceding RWKV-SG module, while the input for each subsequent stage(\( 2 \leq i \leq N \)) originates from the output of the RWKV-PD module in the previous stage. We first use the Feature Extractor to extract the global feature \( f \) from \( F_{i-1} \), and the same Feature Extractor is applied at each stage to extract the relevant hierarchical features. The Feature Extractor consists of PRWKV and the Set Abstraction module. This approach ensures consistency in feature representation. Subsequent to feature extraction, \( P_{i-1} \) and \( f \) are further processed through a PointNet module \cite{Qi2016PointNet} to obtain \( q_i \). We set \( F_{i-1} \) as \( k_i \), and input \( P_{i-1} \), \( q_i \), \( k_i \) into the RWKV-ATTN to obtain the hidden feature \( H_i \). Subsequently, the hidden feature \( H_i \) is passed through the deconvolution module \cite{xiang2022snowflake} to derive \( F_i \). Finally, \( F_i \) is processed by the Rebuild Head and the Segment Head to obtain a displacement vector \( \Delta P \) and the updated semantic label \( L_i \), respectively. The previous position \( P_{i-1} \) is then replicated \( \lambda \) times, and the displacement \( \Delta P \) is added to it, resulting in the updated position \( P_i \).

\subsection{RWKV Attention (RWKV-ATTN)}
\begin{figure}[!t]
  \centering
  \includegraphics[width=0.95\columnwidth]{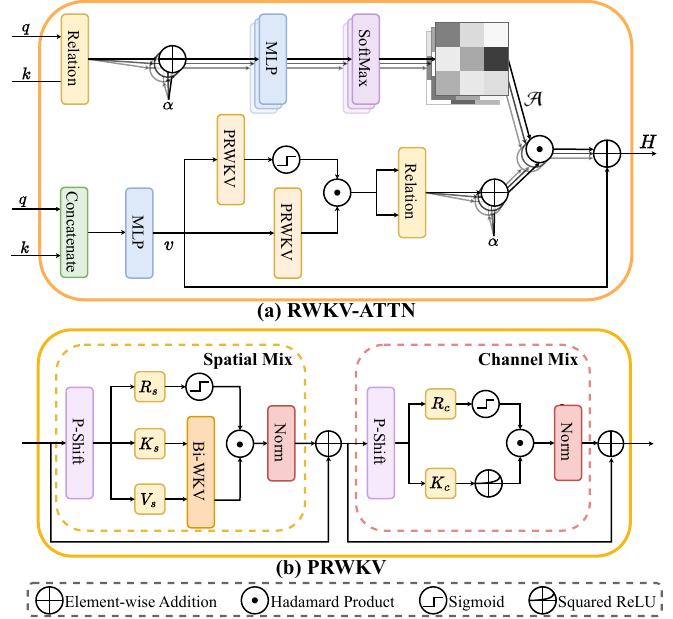}
  \caption{(a) The detailed architecture of the RWKV-ATTN. (b) The detailed architecture of the PRWKV.}
\label{fig:RWKV-ATTN}
\end{figure}

The structure of the RWKV-ATTN module is illustrated in \Cref{fig:RWKV-ATTN} (a). Leveraging the efficient global modeling capability of PRWKV and local attention mechanisms, the proposed RWKV-ATTN module more effectively learns abstract features between parent points and their corresponding child points.

For computing the value \( v \), the query \( q \) and key \( k \) are first concatenated and passed through an MLP to obtain \( v \). The vector \( v \) is fed into two PRWKV modules, and the final value feature \( \hat{v} \) is derived through a gating mechanism, which adaptively controls the global information flow and enhances feature selectivity, as formulated below:
\begin{equation}
    \hat{v} = \sigma(PRWKV(v)) \odot PRWKV(v),
\end{equation}
where \( \sigma \) is the Sigmoid function, \(\odot\) is Hadamard Product.

For computing attention weights, the RWKV-ATTN module employs subtraction relation in a local neighborhood \(L(i)\) (k-nearest neighbors) to calculate local relationships between query \( q \) and key \( k \). These local relationships are then combined with positional encoding \(\alpha \), processed through MLP and SoftMax function to generate the final attention weights \( \mathcal{A} \). Specifically, \( \mathcal{A} \) can be calculated as
\begin{equation}
\mathcal{A} = \frac{exp(MLP(q_i^{}-k_j^{}+\alpha))}{\sum_{j \in L(i)} exp(MLP(q_i^{}-k_j^{}+\alpha))},j \in L(i).
\end{equation}

After obtaining the value \( \hat{v} \) and the attention weights \( \mathcal{A} \), we perform an element-wise multiplication (Hadamard product) between them and then add the intermediate value \( v \) obtained from the MLP. This process can be formally expressed as:
\begin{equation}
    H = \mathcal{A} \odot (\sum_{j \in L(i)}(\hat{v}_i^{}-\hat{v}_j^{}+\alpha)) + v,  j \in L(i).
\end{equation}
This operation leverages the efficient global feature capture of PRWKV alongside local attention mechanisms to extract local features, improving the expressiveness of the hidden features \( H \).

\subsection{Point RWKV (PRWKV)}
An overview of our PRWKV is depicted in \Cref{fig:RWKV-ATTN} (b). PRWKV is designed to efficiently process point clouds by integrating two key components: a spatial-mix module and a channel-mix module. The spatial-mix module functions as an attention mechanism, performing linear complexity global attention computation to capture long-range dependencies within the point cloud. This module is crucial for capturing the global structure of the point cloud. The channel-mix module serves as a feedforward network (FFN), performing feature fusion in the channel dimension to enhance the representation of local features. This dual-module design allows PRWKV to balance global context modeling and local detail preservation effectively.

\begin{figure}[!t]
  \centering
  \includegraphics[width=\columnwidth]{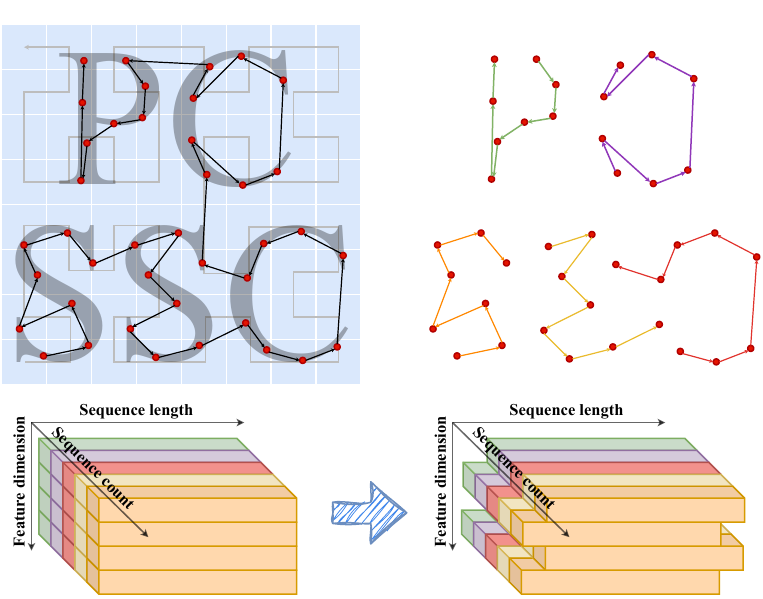}
  \caption{The specific procedure of P-Shift.}
  \label{fig:Point-Shift}
\end{figure}

In the Spatial Mixing module, input features are processed through a shift operation, referred to as \( P\text{-}Shift \), as illustrated in \Cref{fig:Point-Shift}. Unlike images or voxels, point clouds are inherently unordered. To perform the shift operation on point clouds, we first serialize them. Specifically, whenever a new point cloud is introduced, we randomly apply one of the following serialization methods: z-order \cite{1966z-order}, Hilbert-order \cite{hilbert1935stetige}.
This serialization transforms the unordered point cloud into an ordered sequence, enabling the \( P\text{-}Shift \) operation. The \( P\text{-}Shift \) operation is formally defined as:

\begin{equation}
    \begin{aligned}
    &P\text{-Shift}_{(*)}(X) = X + (1 - \mu_{(*)}) X^{'}, \\
    &X^{'} = \text{Concat}(X_1, X_2, X_3, X_4),
    \end{aligned}
\end{equation}
where \( (*) \in \{R, K, V\} \) denotes the interpolation of \( X \) and \( X^{'} \), controlled by the learnable vectors \( \mu_{(*)} \). 

In \( P\text{-Shift} \), the complete point cloud sequence with \( n \) points is uniformly divided into \( s \) subsequences, each containing \( n/s \) points. Thus, \( X^{'} \) represents the slicing vector of \( X \), defined as:\( X \), i.e., \(X_1 = [s, n/s-1, 0 : C/4], \quad X_2 = [s, n/s+1, C/4 : C/2], X_3 = [s, n/s-1, C/2 : 3C/4], \quad X_4 = [s, n/s+1, 3C/4 : C].\)
The \( P\text{-}Shift \) function enhances the attention mechanism by enabling it to focus on proximate points across diverse channels without significantly increasing computational overhead. This process further broadens the receptive field, thereby enhancing the point coverage in subsequent layers. The process of \( P\text{-}Shift \) is shown in \Cref{fig:Point-Shift}.

After \( P\text{-}Shift \), the features projected into matrices \( R_s \), \( K_s \), and \( V_s \in \mathbb{R}^{T \times C} \) via parallel linear transformations:

\begin{equation}
    \begin{aligned}
    R_s &= P\text{-}Shift_R(X)W_R, \\
    K_s &= P\text{-}Shift_K(X)W_K, \\
    V_s &= P\text{-}Shift_V(X)W_V.
    \end{aligned}
\end{equation}

The global attention output $\mathbf{wkv}$ is computed via a linear-complexity bidirectional attention mechanism $Bi\text{-}WKV$ \cite{duan2024vrwkv}, applied to $K_s$ and $V_s$  to enhance contextual modeling:

\begin{equation}
wkv = Bi\text{-}WKV(K_s, V_s).
\end{equation}  

The attention result for the $t$-th feature token is given by:

\begin{equation}
wkv_t = Bi\text{-}WKV(K, V)_t . \\
\frac{\sum_{i=0,i\neq t}^{T-1} e^{-(|t-i|-1)/T \cdot \omega + k_i} v_i + e^{u + k_t} v_t}{\sum_{i=0}^{T-1} e^{-(|t-i|-1)/T \cdot \omega + k_i} + e^{u + k_t}}.
\end{equation}

The output $O_s$ is obtained by element-wise multiplication of $\sigma(R_s)$ and $wkv$, followed by a linear projection and layer normalization:

\begin{equation}
O_s = LayerNorm\left((\sigma(R_s) \odot wkv)W_O\right).
\end{equation}

In the Channel Mixing module, $R_c$ and $K_c$ are obtained similarly, while $V_c$ is computed as a linear projection of the activated $K_c$:
\begin{equation}
    \begin{aligned}
    R_c &= P\text{-}Shift_R(X)W_R, \\
    K_c &= P\text{-}Shift_K(X)W_K, \\
    V_c &= SquaredReLU(K_c)W_V.
    \end{aligned}
\end{equation}

The output $O_c$ is obtained by element-wise multiplication of $\sigma(R_c)$ and $V_c$, followed by a linear projection:
\begin{equation}
O_c = (\sigma(R_c) \odot V_c)W_O.
\end{equation}

\subsection{Training Loss}
For the completion task, we use the Chamfer Distance (CD) to evaluate the difference between the predicted point cloud \( P \) and the ground truth point cloud \( \widehat{P} \).

\begin{equation}
L_{CD} = \frac{1}{|P|} \sum_{x \in P} \min_{y \in \widehat{P}} ||x - y|| + \frac{1}{|\widehat{P}|} \sum_{y \in \widehat{P}} \min_{x \in P} ||y - x||, 
\label{eq:CD}
\end{equation}
where \( x \) and \( y \) are points in \( P \) and \( \widehat{P} \), respectively.

For the semantic segmentation task, we use a weighted negative log-likelihood (NLL) loss to address the challenge of class imbalance. In traditional NLL loss, each class is treated equally, which can lead to suboptimal performance for underrepresented classes. By assigning different weights to each class, we increase the penalty for misclassifying underrepresented classes, thereby encouraging the model to focus more on those classes during training.

The weighted negative log-likelihood loss is defined as:

\begin{equation}
L_{sem} = \sum_{l \in L} - w_{l} \log(p_l),
\label{eq:Seg}
\end{equation}
where \( l \) represents the labels associated with the nearest ground-truth points to the predicted points \( l \in L \), \( p_l \) is the predicted probability, and \( w \) is the weight assigned to each class to mitigate class imbalance. 
In the SSC-PC dataset, \( w \) is set to \{1.50, 0.96, 1.03, 1.10, 1.67, 1.10, 1.14, 1.74, 0.69, 1.06, 2.09, 1.10, 1.06, 1.57, 1.68, 0.69\}.

The overall loss function \( L_{SSC} \) combines both Chamfer Distance loss \( L_{CD} \) and semantic segmentation loss \( L_{sem} \), balanced by the parameter \( \alpha \):

\begin{equation}
L_{SSC} = \sum_{i=0}^{N} L_{CD}(P_i, \widehat{P}_i) + \alpha L_{sem}(L_i, \widehat{L}_i).
\label{eq:SSC}
\end{equation}
In \Cref{eq:SSC}, \( \alpha \) is set to 0.01 for indoor scenes and 0.2 for outdoor scenes due to the larger spatial scale of the latter, \( P_0 \) and \( L_0 \) represent \( P_{coarse} \) and \( L_{coarse} \) respectively. When \( 1 \leq i \leq N \), \( P_i \) and \( L_i \) represent the semantic scene point clouds output by RWKV-PD.

\section{Experiment}
\label{sec:Experiments}

\subsection{Dataset Preparation}

For point cloud semantic scene completion, existing datasets play a crucial role in evaluating model performance across different settings. Specifically, SSC-PC and NYUCAD-PC primarily address indoor environments, whereas PointSSC serves as a valuable benchmark for outdoor scenes. 
In addition to these existing datasets, we introduce two new contributions. First, we present NYUCAD-PC-V2, an improved version of NYUCAD-PC that rectifies semantic ambiguities in object labeling, thereby ensuring more reliable annotations. Second, we introduce 3D-FRONT-PC, a novel dataset derived from 3D-FRONT scene models, which offers a broader spectrum of categories and a substantially larger scale.

\textbf{SSC-PC \cite{zhang2021point,katz2007direct}:} This dataset comprises 1941 scenes from 16 object categories in RGB-D format, accompanied by ground-truth point clouds. Due to the absence of camera intrinsics, partial input point clouds are generated using a visibility approximation method. Both input and output point clouds consist of 4096 points, with 1543 scenes designated for training and 398 for testing.

\textbf{NYUCAD-PC \cite{Xu2023CasFusionNet,firman2016structured,guo2015predicting}:} Comprising 1449 RGB-D images, this dataset includes 795 training and 654 testing samples. The input clouds contain 4096 points, while the ground truth is represented by 8192 points, sampled using Poisson disk sampling. With 11 categories (excluding "empty"), This dataset is more challenging than SSC-PC due to occlusions and incomplete scene coverage.


\textbf{NYUCAD-PC-V2:} Building upon NYUCAD-PC, we introduce NYUCAD-PC-V2, an corrected version of the original dataset. The improvement addresses a significant issue in the original dataset: the semantic confusion caused by adjacent objects being misclassified when they are close to surfaces like walls, floors, or ceilings, which can lead the model to learn incorrect semantic information.

\begin{figure*}[t!]
  \centering
  \includegraphics[width=0.92\textwidth]{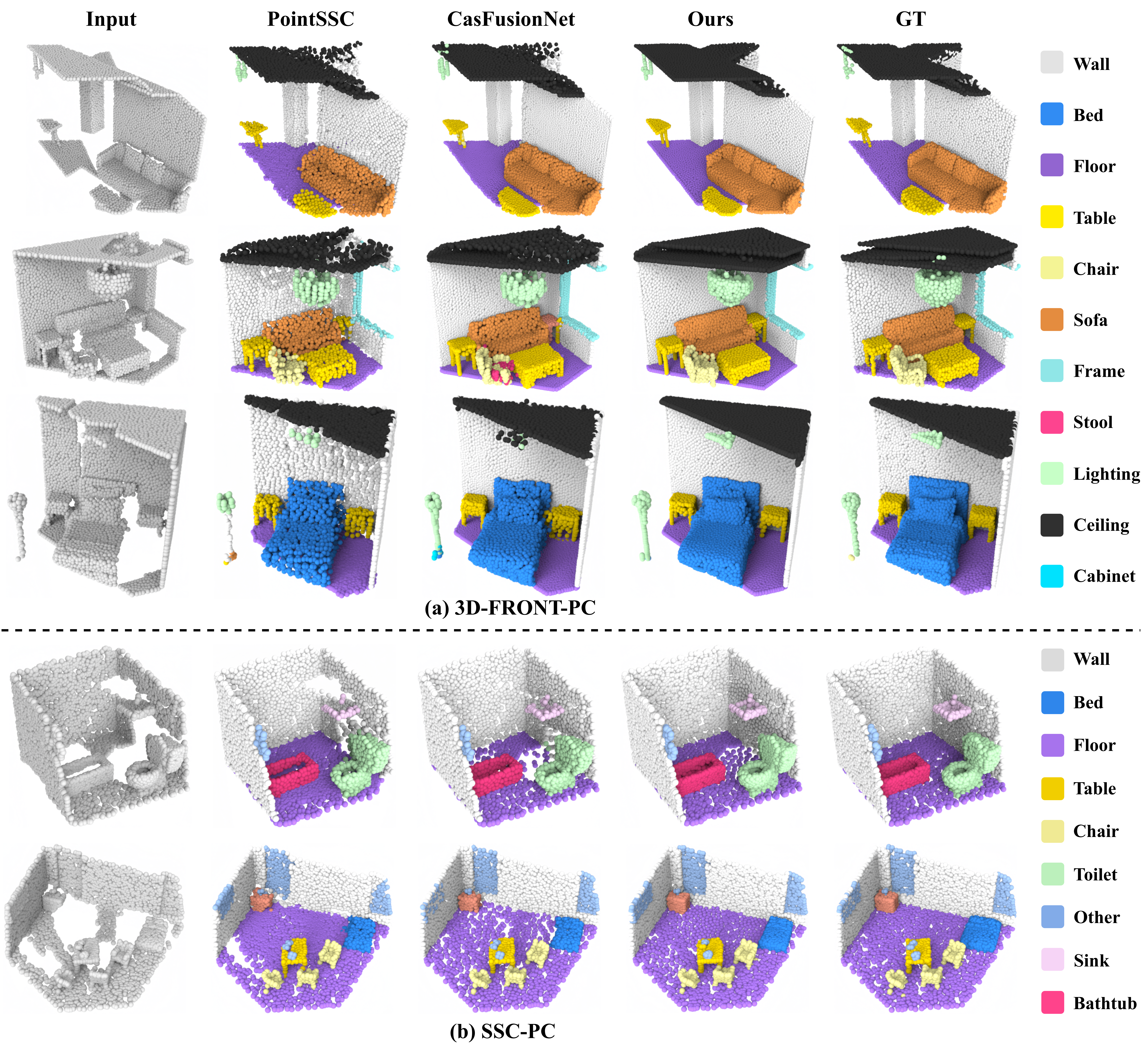}
  \caption{(a) compares point cloud semantic scene completion results on 3D-FRONT-PC, while (b) shows results on SSC-PC.}
  \label{fig:Compare}
\end{figure*}

\textbf{3D-FRONT-PC:} The 3D-FRONT-PC dataset is derived from the 3D-FRONT \cite{fu20213d3d-front} virtual scene dataset provided by Alibaba. It contains 12,655 scene point clouds representing various indoor settings, including 10,124 instances in the training set and 2,531 instances in the test set. 


\textbf{PointSSC:} Derived from six intersections in V2X-Seq, PointSSC comprises 11,275 scene point clouds, with each input point cloud containing 26,624 points and the corresponding ground truth comprising 212,992 points. The dataset is divided based on time and scene. The time-based split allocates 80\% of the frames for training and 20\% for testing, 
while the scene-based split assigns four intersections for training and the remaining two for testing.

\subsection{Evaluation metrics and Implementation} 

\textbf{Evaluation metrics.} For evaluating scene completion performance, we employ Chamfer Distance (CD) and F1-score. For semantic segmentation, we assess performance using the mean class Intersection over Union (mIoU) and the mean class accuracy (mAcc).

\textbf{Implementation.} We implement our network in PyTorch and train it on a single Nvidia RTX 3090 GPU for 100 epochs with a batch size of 8. The AdamMax optimizer is used with an initial learning rate of 0.0075, which decayes by a factor of 0.99 every 2 epochs. The balance parameter $\alpha$ in Eq. \ref{eq:SSC} is set to 0.01. 
Furthermore, taking into account both performance and model complexity, we employ a three-stage RWKV-PD for indoor scenes and a single-stage RWKV-PD for outdoor scenes.

\begin{table*}[t] 
    \begin{center}
    \caption{Quantitative comparison of methods on different indoor datasets. CD refers to Chamfer Distance L2 (multiplied by \boldmath{$10^3$}),F1-score means calculating by distance threshold 0.01. mIoU is the mean Intersection over Union, and mAcc represents mean Accuracy. The best results are in Bold.}
    \label{tab:table_1}
    \resizebox{\textwidth}{!}{
        \begin{tabular}{l|cccc|cccc|cccc|cccc}
            \toprule
            \multirow{2}{*}{Method} & \multicolumn{4}{c|}{SSC-PC} & \multicolumn{4}{c|}{NYUCAD-PC} & \multicolumn{4}{c|}{NYUCAD-PC-V2} & \multicolumn{4}{c}{3D-FRONT-PC} \\
             \cmidrule{2-17}
             & CD $\downarrow$ & F1-Score $\uparrow$ & mAcc $\uparrow$ & mIoU $\uparrow$ & CD $\downarrow$ & F1-Score $\uparrow$ & mAcc $\uparrow$ & mIoU $\uparrow$ & CD$\downarrow$ & F1-Score $\uparrow$ & mAcc $\uparrow$ & mIoU $\uparrow$ & CD $\downarrow$ & F1-Score $\uparrow$ & mAcc $\uparrow$ & mIoU $\uparrow$ \\
            \midrule
            PCSSC-Net\cite{zhang2021point} & 1.58 & - & - & 88.2 & - & - & - & - & - & - & - & - & - & - & - & - \\
            CasFusionNet\cite{Xu2023CasFusionNet} & 0.426 & 54.00\% & 95.11 & 91.94 & 1.164 & 75.19\% & 59.75 & 49.46 & 1.369 & 74.18\% & 60.77 & 50.42 & 0.218 & 84.95\% & 61.31 & 52.69 \\
            PointSSC\cite{yan2024pointssc} & 0.614 & 51.02\% & 88.32 & 82.52 & 2.745 & 65.99\% & 52.05 & 42.12 & 2.698 & 64.96\% & 54.84 & 44.70 & 0.251 & 79.37\% & 61.06 & 55.33 \\            
            \textbf{Ours} & \textbf{0.265} & \textbf{66.93\%} & \textbf{97.99} & \textbf{95.27} & \textbf{1.116} & \textbf{76.30\%} & \textbf{66.47} & \textbf{54.52} & \textbf{1.104} & \textbf{76.77\%} & \textbf{68.49} & \textbf{56.79} & \textbf{0.171} & \textbf{89.21\%} & \textbf{78.33} & \textbf{64.27} \\
            \bottomrule
        \end{tabular}
        }
    \end{center}
\end{table*}

\begin{table}[t] 
    \centering
    \caption{Quantitative comparison of methods on outdoor dataset PointSSC-time-split and PointSSC-scene-split\cite{yan2024pointssc}. CD refers to Chamfer Distance L1 (multiplied by \boldmath{$10^3$}),F1-score means calculating by distance threshold 0.3. The best results are in Bold.}
    \label{tab:table_2}
    \resizebox{0.475\textwidth}{!}{
        \begin{tabular}{l|ccc|ccc}
            \toprule
            \multirow{2}{*}{Method} & \multicolumn{3}{c|}{PointSSC-time-split} & \multicolumn{3}{c}{PointSSC-scene-split} \\
            \cmidrule{2-7}
             & CD $\downarrow$ & F1-Score $\uparrow$ & mIoU $\uparrow$ & CD $\downarrow$ & F1-Score $\uparrow$ & mIoU $\uparrow$ \\
            \midrule
            CasFusionNet\cite{Xu2023CasFusionNet} & 467.76 & 70.84\% & 45.76 & 664.85 & 43.98\% & 11.52 \\
            PointSSC\cite{yan2024pointssc} & 208.94 & 81.42\% & 50.58 & 410.92 & 63.57\% & 14.64 \\            
            \textbf{Ours} & \textbf{179.03} & \textbf{87.72\%} & \textbf{60.30} & \textbf{361.58} & \textbf{64.18\%} & \textbf{17.23} \\ 
            \bottomrule
        \end{tabular}
        }
\end{table}

\subsection{Quantitative and Qualitative Results}
We compared our proposed method with recent point-based semantic scene completion networks, including PCSSC-Net \cite{zhang2021point}, CasFusionNet \cite{Xu2023CasFusionNet}, and PointSSC \cite{yan2024pointssc}, to evaluate its effectiveness and robustness. To ensure a fair and rigorous comparison, we retrained the publicly available CasFusionNet code on our newly prepared NYUCAD-PC-V2 and 3D-FRONT-PC datasets. Additionally, we trained PointSSC using its publicly accessible code on all indoor scenes. 
Notably, for PointSSC-scene-split, PointSSC integrates both point clouds and images as inputs, whereas our method relies solely on point clouds. 
The remaining data were obtained from the published evaluation metrics of PCSSC-Net, CasFusionNet, and PointSSC to ensure objective comparison.

\textbf{Quantitative Results.} As shown in Table \ref{tab:table_1}, we evaluated our method on four indoor datasets: SSC-PC, NYUCAD-PC, NYUCAD-PC-V2 and 3D-FRONT-PC, as well as outdoor datasets: PointSSC-time-split and PointSSC-scene-split, comparing its performance against the indoor baseline CasFusionNet and the outdoor baseline PointSSC. On the SSC-PC dataset, our method reduces the mAcc inaccuracy rate from 4.89\% to 2.01\%, achieving an approximate 58.9\% reduction. Furthermore, the CD L2 metric (multiplied by \(\boldmath{10^3}\)) is reduced from 0.426 to 0.265
.  
In the outdoor scenario of PointSSC-time-split, our method significantly enhances performance, increasing the mIoU from 50.58\% to 60.30\%
. Meanwhile, the CD L1 metric (multiplied by \(\boldmath{10^3}\)) is reduced from 208.94 to 179.03
.


\textbf{Qualitative results.} \Cref{fig:Compare} further presents qualitative comparisons on two datasets. \Cref{fig:Compare} (a) shows three scenes from the 3D-FRONT-PC dataset, while \Cref{fig:Compare} (b) displays two scenes from the SSC-PC \cite{zhang2021point} dataset. From the visual comparisons, our method exhibits superior performance in both segmentation and completion tasks, effectively preserving and reconstructing fine details. 

\subsection{Ablation Study}

To evaluate the effectiveness of the major components in our method, we conducted an ablation study on the SSC-PC dataset by simplifying our network under the following configurations:


\begin{description}[align=left]
  \item[Model A:] We remove the RWKV-SG module, and adopt CasFusionNet \cite{Xu2023CasFusionNet} approach that utilizes global features to predict semantic labels with MLPs and applies FPS to obtain a coarse point cloud.
  \item[Model B:] We combine the Snowflake Point Deconvolution (SPD) module\cite{xiang2022snowflake} with our Segment Head and Rebuild Head to replace the RWKV-PD module.
  \item[Model C:] We set the number of layers in the RWKV-PD module to 1 layer, reducing the complexity of the model and evaluating the performance with fewer layers.
\end{description}

\begin{table}[t!]
    \centering
    \caption{Comparison of different models for scene completion and semantic segmentation.}
    \label{tab:table_3}
    \begin{tabular}{l|c|cc}
        \toprule
        \multirow{2}{*}{Model} & \multicolumn{1}{c|}{Scene completion} & \multicolumn{2}{c}{Semantic segmentation} \\
        \cmidrule{2-4}
        ~ & CD ($\times10^{-3}$) & mAcc & mIoU \\
        \midrule
        A & 0.353 & 97.49 & 94.13 \\
        B & 0.274 & 97.26 & 94.61 \\
        C & 0.287 & 97.72 & 94.48 \\
        \cmidrule{1-4}
        \textbf{Ours} & \textbf{0.265} & \textbf{97.99} & \textbf{95.27} \\
        \bottomrule
    \end{tabular}
\end{table}

Subsequent to the ablation study, we observe that each modification to the model leads to distinct changes in performance. 


Specifically, Model A, which replaces the RWKV-SG module with MLPs and FPS, results in a notable increase in Chamfer Distance. This demonstrates that the coarse point cloud and point-wise features generated by RWKV-SG significantly contribute to the subsequent coarse-to-fine process.
Model B, after replacing the RWKV-PD module, failed to effectively capture finer spatial and sementic details, leading to lower performance in both scene completion and segmentation accuracy.
Model C, which simplifies the RWKV-PD module by reducing its number of layers to just one, leads to a performance drop across both scene completion and semantic segmentation tasks. 
This indicates that the additional layers in the RWKV-PD module are essential for capturing the complexity of point-wise features.

\section{Conclusion}
\label{sec:Conclusion}

In this work, we introduce RWKV-PCSSC, a lightweight point cloud semantic scene completion network inspired by the RWKV mechanism. The network comprises two key components: the RWKV Seed Generator (RWKV-SG), which aggregates features from partial point clouds to generate a coarse representation, and the RWKV Point Deconvolution (RWKV-PD) modules, which progressively restore point-wise features. Our method achieves state-of-the-art performance on both indoor and outdoor datasets, while significantly reducing the number of parameters and memory usage. 








\section{Dataset Details}
\textbf{NYUCAD-PC-V2:} We present NYUCAD-PC-V2, an enhanced version of the NYUCAD-PC dataset that addresses semantic misclassifications of adjacent objects near surfaces such as walls, floors, and ceilings, as shown in Figure 6. 
To resolve this, we reassign points located within 0.005 distance of walls, floors, and ceilings to the corresponding mesh label of the nearest surface. This refinement leads to a more accurate semantic segmentation, where surfaces are better distinguished from attached objects.

\textbf{3D-FRONT-PC Dataset:} The 3D-FRONT-PC dataset is derived from the Alibaba-provided 3D-FRONT virtual scene dataset~\cite{fu20213d3d-front}, comprising 12,655 indoor scene point clouds with diverse settings. It consists of 10,124 training instances and 2,531 test instances.

\begin{figure}[!t]
  \centering
  \includegraphics[width=\columnwidth]{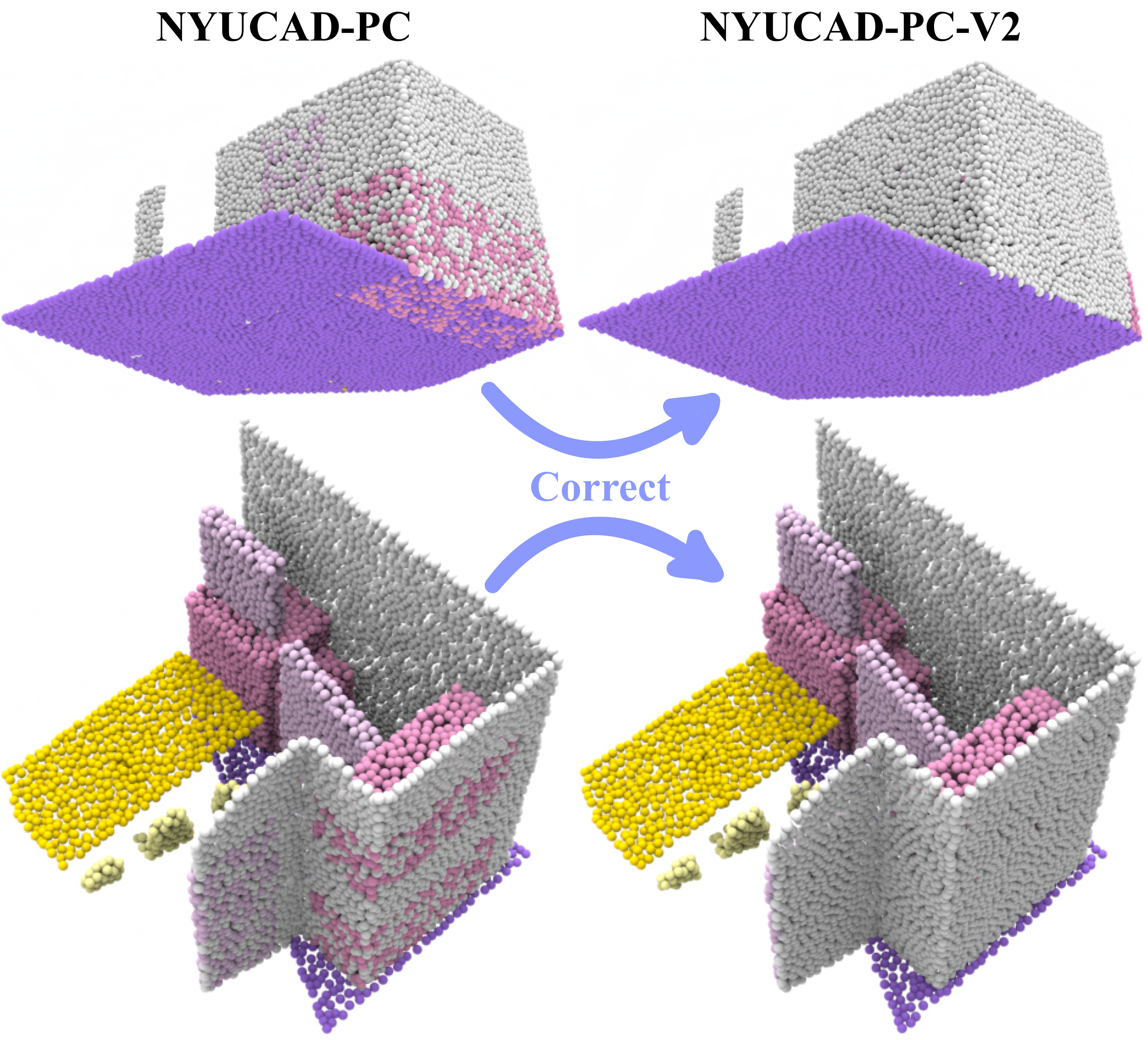}
  \caption{The issue of label confusion in the point clouds within the NYUCAD-PC dataset, along with the corrections made in the NYUCAD-PC-V2 dataset.}
\label{fig:NYUCAD_Correct}
\end{figure}

To generate the input point cloud \( P_{in} \), we employ raycasting to render three-dimensional scenes into two-dimensional depth images via a pinhole camera model. Specifically, we define the intrinsic parameters (focal length and optical center) and extrinsic parameters (position, orientation, and field of view) of the camera, and set the output image resolution. Based on these parameters, rays are generated from the camera's viewpoint, traversing each pixel of the image plane and simulating intersections with the scene. These rays are cast into the three-dimensional scene, and the distances of their first intersections with the scene surfaces are recorded and organized into a depth image, where each pixel value represents the Euclidean distance from the camera to the scene surface. The depth information is then converted into a point cloud by integrating the depth image with the camera's intrinsic parameters and back-projecting each pixel based on its depth value, resulting in \( P_in \) that encapsulates geometric information.

To acquire the ground truth points \( P_{gt} \) within the camera's field of view (FoV), we first generate a dense scene point cloud \( P_{scene} \) through Poisson sampling on the 3D mesh representation of the environment. The sampling density is configured to ensure uniform coverage of surface geometries. We subsequently perform frustum culling on \( P_{scene} \) using a near clipping plane at \( 0.2 \) m, far clipping plane at \( 6 \) m, and \( 60^\circ \) FoV. Points satisfying both the depth constraint and the angular bounds are retained as candidates for visible ground truth. Finally, Farthest Point Sampling (FPS) is applied to \( P_{scene} \) and the visible subset to obtain downsampled point clouds of \( 4096 \) and \( 8192 \) points respectively.


\section{Additional Notes on Modules}

The detailed architectures of the Feature Extractor and the Segment Head are depicted in Figure 6 (a) and Figure 6 (b), respectively. The Feature Extractor utilizes a PRWKV module to capture and integrate global contextual information, followed by a series of Set Abstraction layers to progressively refine local point features, culminating in a comprehensive global feature representation, denoted as $f$. On the other hand, the Segment Head employs KPConvD\cite{Thomas2024KPConvX} with residual connections to extract local features from the point cloud, thereby improving feature stability and alleviating gradient vanishing. It then leverages the PRWKV module to abstract global information, effectively modeling long-range dependencies and contextual relationships within the point cloud. Finally, the processed features are fed into an MLP to generate the semantic prediction outputs $L$.



\section{Additional Charts and Results}
Due to page limitations, we present only the overall performances in the main text. For comprehensive results, detailed metrics for different models across the PointSSC, SSC-PC, NYUCAD-PC, NYUCAD-PC-V2, and 3D-FRONT-PC datasets are provided in the supplementary material, as demonstrated in Table 4, Table 5, Table 6, Table 7, and Table 8. Figure 8 provides the qualitative comparisons on PointSSC, and Figure 9 illustrates the qualitative comparisons on NYUCAD-PC-V2.


\begin{figure}[!t]
  \centering
  \includegraphics[width=\columnwidth]{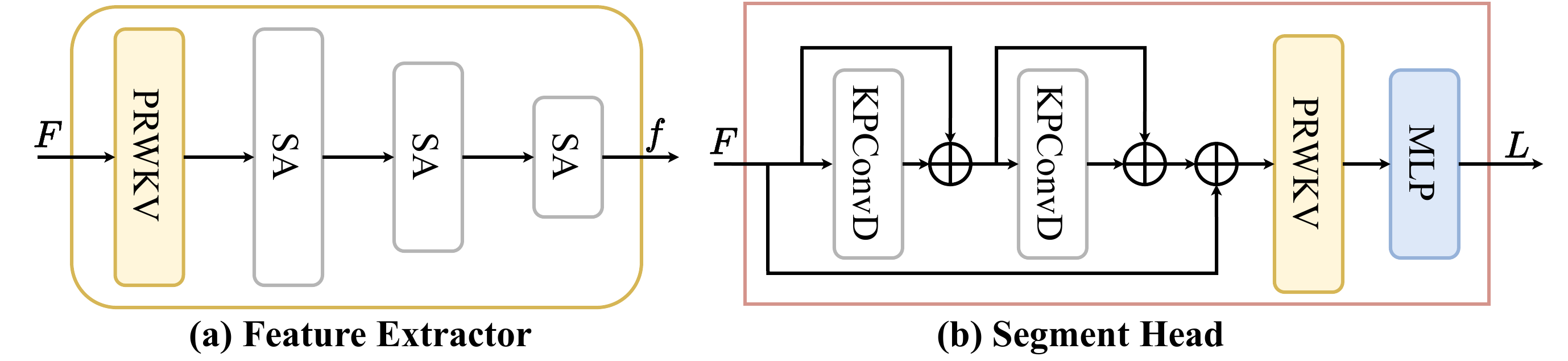}
  \caption{(a) The detailed architecture of the Feature Extractor. (b) The detailed architecture of the Segment Head.}
\label{fig:FE_SH}
\end{figure}

\begin{table*}[t!]
  \centering
  \caption{Quantitative comparison of different methods on PointSSC dataset. CD Means Chamfer Distance(Multiplied By 1000), lower is better. F1-score@0.3 means calculating by distance threshold 0.3. The best results are in \textbf{Bold}. Additionally, PointSSC utilizes both point clouds and images as inputs for Scene Splitting, whereas our method relies solely on point clouds.}
  \vspace{-1pt}  
  \resizebox{\linewidth}{!}{
    \begin{tabular}{c|l|ccc|ccccccccccccccccc}
    \toprule
         & Methods & CD ($L_1$) $\downarrow$  & CD ($L_2$) $\downarrow$ & \multicolumn{1}{l|}{F1-Score@0.3 $\uparrow$} & mIoU $\uparrow$ & building & tree & road & sidewalk & person & plant & car & fence & signboard & bus & truck & streetlight & barricade & van & bicycle & motorcyclist  \\
    \midrule
        \multirow{9}[4]{*}{\begin{sideways}\rotatebox{0}{Time Splitting}\end{sideways}} 
        & FoldingNet\cite{yang2018foldingnet} & 610.30 & 2785.44 & 39.65\% & 14.91 & 8.56 & 60.74 & 81.01 & 27.45 & 0.00 & 35.26 & 12.31 & 11.24 & 1.07 & 0.14 & 0.70 & 0.00 & 0.00 & 0.00 & 0.00 & 0.00 \\
        & PCN\cite{yuan2018pcn} & 317.38 & 510.19 & 64.46\% & 12.65 & 0.14 & 69.71 & 83.91 & 0.70 & 0.00 & 29.76 & 2.08 & 9.05 & 7.05 & 0.01 & 0.00 & 0.00 & 0.01 & 0.00 & 0.00 & 0.00 \\
        & PoinTr\cite{yu2021pointr} & 3864.43 & 129947.08 & 15.66\% & 5.21 & 0.00 & 2.69 & 0.00 & 0.00 & 0.00 & 0.00 & 2.52 & 0.00 & 0.00 & 0.00 & 0.00 & 0.00 & 0.00 & 0.00 & 0.00 & 0.00 \\
        & SnowFlakeNet\cite{xiang2022snowflake} & 461.89 & 2846.09 & 52.99\% & 20.78 & 11.58 & 80.65 & 80.79 & 32.47 & 0.00 & 48.84 & 1.47 & 35.63 & 36.34 & 2.03 & 1.47 & 1.14 & 0.00 & 0.00 & 0.00 & 0.00 \\
        & PMP-Net++\cite{wen2022pmp} & 541.82 & 3753.60 & 57.09\% & 25.25 & 21.79 & 80.67 & 87.89 & 10.10 & 4.73 & 37.53 & 44.29 & 14.53 & 7.74 & 37.97 & 29.66 & 3.20 & 19.02 & 3.83 & 0.00 & 1.04 \\
        & CasFusionNet\cite{Xu2023CasFusionNet} & 467.76 & 5087.18 & 70.84\% & 45.76 & 55.20 & 90.50 & 90.11 & 55.08 & 19.52 & 69.91 & 59.38 & 58.46 & 45.35 & 47.89 & 37.62 & 12.53 & 9.66 & 39.85 & 13.80 & 27.37 \\
        & AdaPoinTr\cite{Yu2023AdaPoinTr} & 237.11 & 290.71 & 78.79\% & 45.09 & 45.53 & 87.08 & \textbf{93.42} & 53.36 & 13.00 & 64.46 & 55.99 & 56.41 & 42.36 & 56.44 & 37.56 & 15.79 & 17.61 & 43.05 & 11.31 & 28.07 \\
        & PointSSC\cite{yan2024pointssc} & 208.94 & 248.28 & 81.42\% & 50.58 & 61.42 & 90.94 & 92.41 & 63.32 & 13.40 & 74.33 & 66.17 & 67.53 & 47.93 & 61.37 & 44.35 & 14.92 & 18.82 & 43.69 & 17.64 & 31.07 \\
        \cmidrule{2-22} & RWKV-PCSSC (Ours) & \textbf{179.026} & \textbf{166.346} & \textbf{87.72\%} & \textbf{60.30} & \textbf{66.45} & \textbf{92.48} & 91.46 & \textbf{65.66} & \textbf{38.04} & \textbf{76.59} & \textbf{67.6} & \textbf{68.4} & \textbf{53.6} & \textbf{68.56} & \textbf{58.69} & \textbf{30.95} & \textbf{29.17} & \textbf{56.86} & \textbf{40.29} & \textbf{60.05} \\
    \midrule
    \multirow{9}[4]{*}{\begin{sideways}\rotatebox{0}{Scene Splitting}\end{sideways}} 
        & FoldingNet\cite{yang2018foldingnet} & 1146.07 & 16950.85 & 6.06\% & 8.11 & 0.14 & 8.79 & 84.06 & 4.08 & 0.00 & 31.60 & 1.01 & 0.00 & 0.00 & 0.00 & 0.00 & 0.00 & 0.00 & 0.00 & 0.00 & 0.00 \\
        & PCN\cite{yuan2018pcn} & 809.55 & 5778.96 & 42.66\% & 8.62 & 0.03 & 47.95 & 78.07 & 0.01 & 0.00 & 11.86 & 0.00 & 0.00 & 0.02 & 0.00 & 0.00 & 0.00 & 0.00 & 0.00 & 0.00 & 0.00 \\
        & PoinTr\cite{yu2021pointr} & 4164.05 & 90275.69 & 15.89\% & 0.40 & 0.00 & 0.00 & 0.02 & 0.01 & 0.00 & 0.00 & 6.06 & 0.00 & 0.00 & 0.00 & 0.24 & 0.00 & 0.00 & 0.00 & 0.00 & 0.00 \\
        & SnowFlakeNet\cite{xiang2022snowflake} & 1055.09 & 21849.79 & 42.36\% & 8.09 & 0.00 & 55.81 & 63.33 & 2.06 & 0.00 & 7.51 & 0.00 & 0.00 & 0.26 & 0.18 & 0.00 & 0.35 & 0.00 & 0.00 & 0.00 & 0.00 \\
        & PMP-Net++\cite{wen2022pmp} & 530.80 & 3753.60 & 57.09\% & 13.66 & 0.00 & 65.75 & 84.76 & 8.80 & 0.00 & 16.32 & 24.66 & 0.01 & 0.86 & 4.96 & 5.17 & 0.00 & \textbf{0.71} & 4.94 & 0.00 & 1.57 \\
        & CasFusionNet\cite{Xu2023CasFusionNet} & 664.85 & 8310.55 & 43.98\% & 11.52 & 0.52 & 62.33 & 65.68 & 6.04 & 0.08 & 9.06 & 21.84 & 0.00 & 0.48 & 8.08 & 0.12 & 0.18 & 0.00 & 7.06 & 0.92 & 1.85 \\
        & AdaPoinTr\cite{Yu2023AdaPoinTr} & 493.41 & 3098.48 & 61.41\% & 13.20 & 0.43 & 66.38 & 84.91 & 5.57 & 0.00 & 15.86 & 15.67 & 0.00 & 0.21 & 1.90 & 5.07 & \textbf{1.55} & 0.00 & 11.55 & 0.06 & 2.06 \\
        & PointSSC\cite{yan2024pointssc} & 410.92 & \textbf{1413.60} & 63.57\% & 14.64 & 0.12 & 68.93 & \textbf{85.68} & \textbf{14.02} & 0.00 & 13.22 & 24.69 & \textbf{10.64} & 1.78 & 1.32 & \textbf{5.65} & 0.01 & 0.07 & 5.42 & 0.54 & 2.13 \\
        \cmidrule{2-22} & RWKV-PCSSC (Ours) & \textbf{361.58} & 1560.67 & \textbf{64.18\%} & \textbf{17.23} & \textbf{1.20} & \textbf{60.24} & 79.79 & 10.18 & \textbf{2.25} & \textbf{20.61} & \textbf{38.13} & 0.09 & \textbf{11.24} & \textbf{8.89} & 1.03 & 0.10 & 0.00 & \textbf{23.58} & \textbf{6.89} & \textbf{14.53} \\		

    \bottomrule
    \end{tabular}%
    }
  \vspace{8pt}  
  \label{tab:pointssc}%
\end{table*}%


\begin{table*}[t!]
  \centering
  \caption{Quantitative comparison of different methods on SSC-PC dataset. CD Means Chamfer Distance(Multiplied By 1000), lower is better. F1-score@0.01 means calculating by distance threshold 0.01. The best results are in \textbf{Bold}. In the original PCSSC-Net paper \cite{liang2021sscnav}, IoU was reported for only 13 categories in the SSC-PC dataset, while the remaining three categories were denoted as "-".}
  \vspace{-1pt}  
  \resizebox{\linewidth}{!}{
    \begin{tabular}{l|ccc|cccccccccccccccccc}
    \toprule
         Methods & CD ($L_1$) $\downarrow$  & CD ($L_2$) $\downarrow$ & \multicolumn{1}{l|}{F1-Score@0.01 $\uparrow$} & mIoU $\uparrow$ & Bathtub & bed & bookshelf & cabinet & ceiling & chair & desk & door & floor & other & sink & sofa & table & toilet & tv & wall  \\
    \midrule
        PCSSC-Net\cite{liang2021sscnav} & - & 1.58 & - & 88.2 & 90.5 & 96.9 & 84.9 & 79.0 & - & 83.7 & 87.4 & 89.1 & 99.1 & - & 89.2 & 87.7 & 86.0 & 86.5 & - & 97.5 \\
        CasFusionNet\cite{Xu2023CasFusionNet} & 9.413 & 0.425 & 54.00\% & 91.94 & 90.55 & 96.81 & 94.55 & 89.92 & 96.43 & 92.80 & 87.44 & 87.64 & 93.17 & 73.38 & \textbf{99.62} & 91.32 & 91.71 & 93.70 & 95.57 & 96.50 \\
        PointSSC\cite{yan2024pointssc} & 11.543 & 0.614 & 51.02\% & 82.52 & 73.06 & 90.22 & 78.62 & 80.05 & 92.29 & 79.88 & 77.04 & 73.45 & 91.02 & 57.29 & 99.29 & 80.47 & 85.11 & 79.95 & 93.33 & 89.30 \\
        \cmidrule{1-21} RWKV-PCSSC (Ours) & \textbf{6.937} & \textbf{0.265} & \textbf{66.93\%} & \textbf{95.27} & \textbf{95.77} & \textbf{97.66} & \textbf{96.70} & \textbf{95.63} & \textbf{97.28} & \textbf{95.72} & \textbf{95.33} & \textbf{93.87} & \textbf{93.46} & \textbf{81.68} & 99.38 & \textbf{95.30} & \textbf{95.21} & \textbf{96.35} & \textbf{98.03} & \textbf{96.93} \\
    \bottomrule
    \end{tabular}%
    }
  \vspace{8pt}  
  \label{tab:ssc_pc}%
\end{table*}%

\begin{table*}[t!]
  \centering
  \caption{Quantitative comparison of different methods on NYUCAD-PC dataset. CD Means Chamfer Distance(Multiplied By 1000), lower is better. F1-score@0.01 means calculating by distance threshold 0.01. The best results are in \textbf{Bold}.}
  \vspace{-1pt}  
  \resizebox{\linewidth}{!}{
    \begin{tabular}{l|ccc|cccccccccccc}
    \toprule
         Methods & CD ($L_1$) $\downarrow$  & CD ($L_2$) $\downarrow$ & \multicolumn{1}{l|}{F1-Score@0.01 $\uparrow$} & mIoU $\uparrow$ & ceiling & floor & wall & window & chair & bed & sofa & table & tvs & furnsink & objs \\
    \midrule
        CasFusionNet\cite{Xu2023CasFusionNet} & 10.276 & 1.164 & 75.19\% & 49.46 & 76.48 & 87.35 & 67.33 & 8.40 & 62.74 & 52.73 & 47.26 & 48.90 & 14.30 & 44.04 & 34.57\\
        PointSSC\cite{yan2024pointssc} & 13.898 & 2.745 & 65.99\% & 42.12 & 80.40 & 84.32 & 60.58 & 6.83 & 43.67 & 44.28 & 37.39 & 35.26 & 8.69 & 37.59 & 24.24 \\
        \cmidrule{1-16} RWKV-PCSSC (Ours) & \textbf{10.128} & \textbf{1.116} & \textbf{76.30\%} & \textbf{54.52} & \textbf{81.22} & \textbf{87.71} & \textbf{70.60} & \textbf{20.29} & \textbf{66.83} & \textbf{61.78} & \textbf{57.94} & \textbf{53.13} & \textbf{15.92} & \textbf{46.89} & \textbf{37.37} \\
    \bottomrule
    \end{tabular}%
    }
  \vspace{8pt}  
  \label{tab:nyucad_pc}%
\end{table*}%

\begin{table*}[t!]
  \centering
  \caption{Quantitative comparison of different methods on NYUCAD-PC-V2 dataset. CD Means Chamfer Distance(Multiplied By 1000), lower is better. F1-score@0.01 means calculating by distance threshold 0.01. The best results are in \textbf{Bold}.}
  \vspace{-1pt}  
  \resizebox{\linewidth}{!}{
    \begin{tabular}{l|ccc|cccccccccccc}
    \toprule
         Methods & CD ($L_1$) $\downarrow$  & CD ($L_2$) $\downarrow$ & \multicolumn{1}{l|}{F1-Score@0.01 $\uparrow$} & mIoU $\uparrow$ & ceiling & floor & wall & window & chair & bed & sofa & table & tvs & furnsink & objs \\
    \midrule
        CasFusionNet\cite{Xu2023CasFusionNet} & 10.410 & 1.369 & 74.18 & 50.42 & 80.35 & 96.29 & 71.29 & 10.45 & 63.80 & 56.21 & 42.01 & 48.02 & \textbf{10.63} & 44.44 & 31.17\\
        PointSSC\cite{yan2024pointssc} & 14.131 & 2.698 & 64.9550 & 44.70 & 77.70 & 95.84 & 67.72 & 4.35 & 38.85 & 55.98 & 37.42 & 36.83 & 10.56 & 42.46 & 24.03 \\
        \cmidrule{1-16} RWKV-PCSSC (Ours) & \textbf{10.126} & \textbf{1.104} & \textbf{76.767} & \textbf{56.79} & \textbf{80.94} & \textbf{96.27} & \textbf{76.96} & \textbf{18.36} & \textbf{69.41} & \textbf{67.64} & \textbf{58.08} & \textbf{55.31} & 10.60 & \textbf{51.42} & \textbf{39.71} \\
    \bottomrule
    \end{tabular}%
    }
  \vspace{8pt}  
  \label{tab:nyucad_pc_v2}%
\end{table*}%

\begin{table*}[t!]
  \centering
  \caption{Quantitative comparison of different methods on 3D-FRONT-PC dataset. CD Means Chamfer Distance(Multiplied By 1000), lower is better. F1-score@0.01 means calculating by distance threshold 0.01. The best results are in \textbf{Bold}.}
  \vspace{-1pt}  
  \resizebox{\linewidth}{!}{
    \begin{tabular}{l|ccc|ccccccccccccccccccccc}
    \toprule
         Methods & CD ($L_1$) $\downarrow$  & CD ($L_2$) $\downarrow$ & \multicolumn{1}{l|}{F1-Score@0.01 $\uparrow$} & mIoU $\uparrow$ & cabinet & pier & bed & chair & table & sofa & lighting & storage & electric & decor & sink & attachment & plants & wall & ceiling & floor & openings & pipe & stair & others \\
    \midrule
        CasFusionNet\cite{Xu2023CasFusionNet} & 6.947 & 0.218 & 84.95 & 52.69 & 46.19 & 22.01 & 76.18 & 75.48 & 74.32 & 79.38 & 85.83 & 44.07 & 39.43 & 21.83 & 45.45 & 22.46 & 45.99 & 84.20 & 90.31 & \textbf{92.85} & 69.97 & \textbf{36.14} & 1.50 & 40.81\\
        PointSSC\cite{yan2024pointssc} & 7.627 & 0.251 & 79.368 & 55.33 & 41.60 & 24.49 & 80.22 & 79.55 & 81.02 & 83.29 & 90.70 & 37.57 & 30.06 & 18.64 & 37.22 & 1.66 & 48.96 & 87.88 & 86.95 & 91.37 & 71.44 & 31.02 & 0.00 & 30.06 \\
        \cmidrule{1-25} RWKV-PCSSC (Ours) & \textbf{6.421} & \textbf{0.171} & \textbf{89.21} & \textbf{64.27} & \textbf{55.57} & \textbf{53.95} & \textbf{88.63} & \textbf{86.54} & \textbf{86.88} & \textbf{89.35} & \textbf{92.15} & \textbf{54.33} & \textbf{46.21} & \textbf{39.15} & \textbf{64.66} & \textbf{50.42} & \textbf{66.06} & \textbf{88.74} & \textbf{90.83} & 91.12 & \textbf{77.59} & 7.81 & \textbf{4.85} & \textbf{50.52}  \\
    \bottomrule
    \end{tabular}%
    }
  \vspace{8pt}  
  \label{tab:3d_front_pc}%
\end{table*}%

\cleardoublepage

\begin{figure*}[p]
  \centering
  \includegraphics[width=\textwidth]{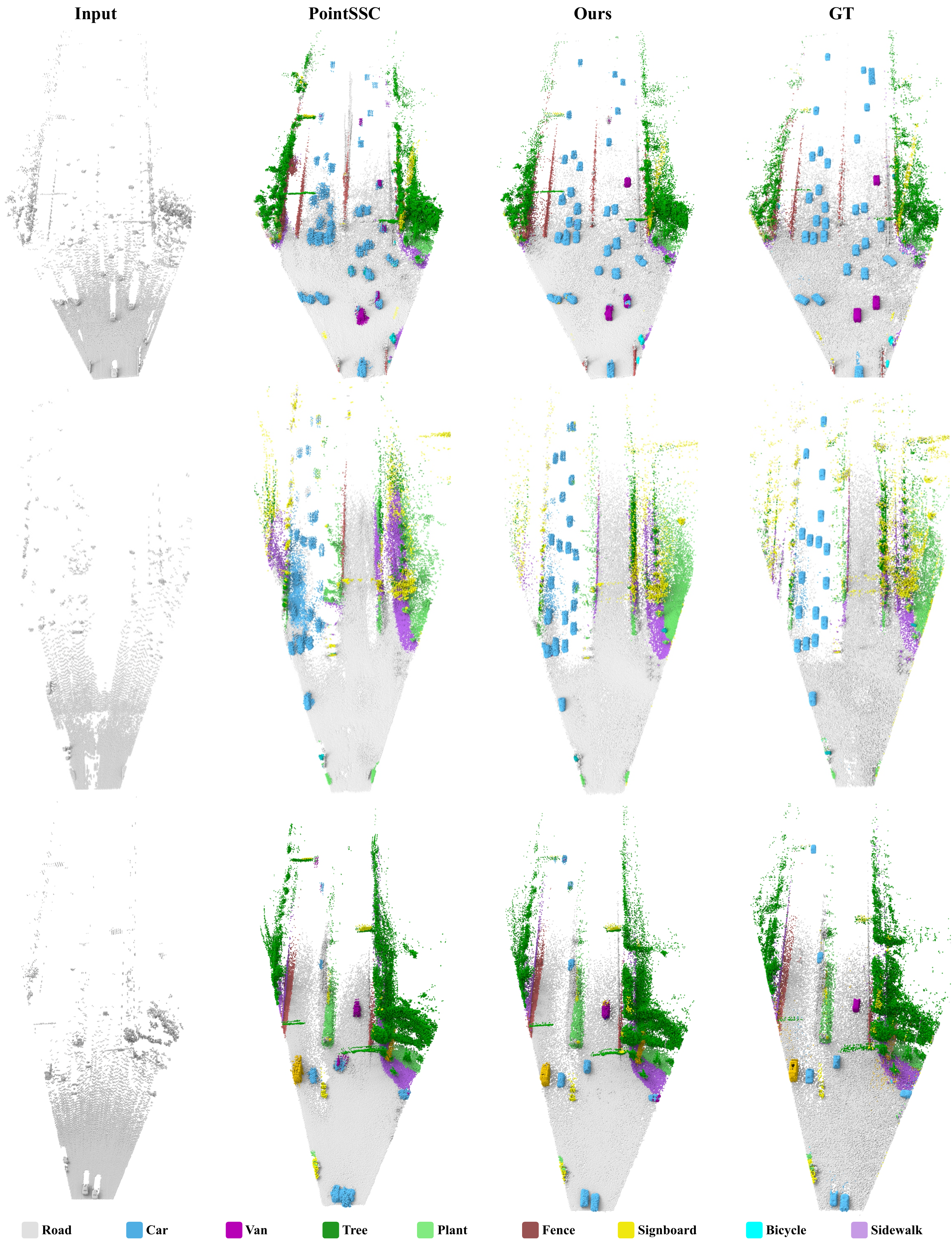}
  \caption{Visualization results on the PointSSC dataset}
  \label{fig:/PointSSC-Compare}
\end{figure*}
\FloatBarrier 








\appendix

\begin{acks}
The work is supported by National Key Research and Development Program of China (2025YFB3003600), National Natural Science Foundation of China (No.62202151) and (No.62202152) .
\end{acks}

\bibliographystyle{ACM-Reference-Format}
\bibliography{sample-base}









\end{document}